\documentclass[a4paper]{article}

\usepackage{INTERSPEECH2016}
\usepackage{graphicx}
\usepackage{amssymb,amsmath,bm}
\usepackage{textcomp}

\usepackage{amsmath,graphicx}
\usepackage{url}
\usepackage{mathtools}
\usepackage{multicol,multirow}
\usepackage{caption}
\usepackage{subfigure,ctable}
\usepackage{enumerate} 

\newcommand{\myfigureshrinker}{\vspace{-0.25cm}}

\ninept

\title{Generation and Pruning of Pronunciation Variants to Improve ASR Accuracy}

\makeatletter
\def\name#1{\gdef\@name{#1\\}}
\makeatother \name{{\em Zhenhao Ge, Aravind Ganapathiraju, Ananth N. Iyer, Scott A. Randal, Felix I. Wyss}}

\address{Interactive Intelligence Inc., Indianapolis, Indiana, USA \\
  {\small \tt \{roger.ge, aravind.ganapathiraju, ananth.iyer, scott.randal, felix.wyss\}@inin.com}
}

\begin{document}

  \maketitle
  \begin{abstract}
   Speech recognition, especially name recognition, is widely used in phone services such as company directory dialers, stock quote providers or location finders. It is usually challenging due to pronunciation variations. This paper proposes an efficient and robust data-driven technique which automatically learns acceptable word pronunciations and updates the pronunciation dictionary to build a better lexicon without affecting recognition of other words similar to the target word. It generalizes well on datasets with various sizes, and reduces the error rate on a database with 13000+ human names by 42\%, compared to a baseline with regular dictionaries already covering canonical pronunciations of 97\%+ words in names, plus a well-trained spelling-to-pronunciation (STP) engine.
  \end{abstract}

\section{Introduction}
\label{sec:intro}

Grammar-based Automatic Speech Recognition (ASR), can be challenging due to variation of pronunciations. These variations can be pronunciations of native words from non-natives, or pronunciations of imported non-native words from natives, or it may be caused by uncommon spelling of some special words. Many techniques have been tried to address this challenge, such as weighted speaker clustering, massive adaptation, and adaptive pronunciation modeling \cite{gao2001innovative}.

Words specified in the grammar have their baseline pronunciations either covered in regular dictionaries, such as 1) prototype dictionaries for the most common words, and 2) linguist hand-crafted dictionaries for the less common words, or 3) generated using a spelling-to-pronunciation (STP)/grapheme-to-phoneme (G2P) engine with a set of rules for special words. These baseline pronunciations sometimes have a big mismatch for the ``difficult'' words and may cause recognition errors, and ASR accuracy can be significantly improved if more suitable pronunciations can be learned. However, blindly generating variants of word pronunciation, though it can increase the recognition rate for that particular word, will reduce accuracy recognizing potential ``similar'' words, which are close to the target word in the pronunciation space.

There are various ways to learn pronunciations \cite{badr2011pronunciation, chan2012discriminative} and here we propose a novel efficient algorithm. The goal of this algorithm is two-fold. For each target word: a) select the best set of alternate pronunciations from a candidate set originating from the baseline pronunciation; b) avoid any ``side-effect'' on neighboring words in pronunciation space. This is achieved by maximizing the overall recognition accuracy of a word set containing the target word and its neighboring words. A pronunciation variant generation and searching process is developed, which further performs sorting and pruning to limit the number of total accepted pronunciations for each word. 

Beaufays et al. used probability models to suggest alternative pronunciations by changing one phoneme at a time \cite{beaufays2003learning}. Reveil et al. adds pronunciation variants to a baseline lexicon using multiple phoneme-to-phoneme (P2P) converters with different features and rules  \cite{reveil2012improving}. Compared to these methods, the proposed technique is more efficient and allows searching in a much wider space without affecting accuracy. 

The work was initiated during the first author's internship at Interactive Intelligence (ININ) \cite{ge2013mispronunciation}, and later improved in terms of accuracy, efficiency and flexibility. 
This paper is organized as follows: Sec. \ref{sec:data} describes the database, Sec. \ref{sec:overview} provides an overview of the grammar-based name recognition framework; Sec. \ref{sec:preliminary} introduces some preliminary knowledge which faciliates the explanation of the pronunciation learning algorithm in Sec. \ref{sec:pronlearn}; Sec. \ref{sec:improvement} provides some heuristics to improve efficiency in implementing pronunciation learning, followed by the results and conclusions in Sec. \ref{sec:results}.

\section{Data}
\label{sec:data}
This work used the ININ company directory database, which contains human names (concatenation of 2,3, or 4 words), intentionally collected from 2 phases for pronunciation learning (training) and accuracy improvement evaluation (testing) respectively. They share the same pool of 13875 names, and Tab. \ref{tab:datasets} lists the statistics and baseline accuracies. Names were pronounced in English by speakers from multiple regions and countries. They were asked to read a list of native and non-native names with random repetitions. Then, the audio was segmented into recordings of individual names. The reduction in Name Error Rate (NER) from phase 1 to phase 2 was mainly because the latter were recorded in cleaner channels with less packet loss, and better corpus creation methods.
\begin{table}[tb]
\captionsetup[table]{aboveskip=0pt}
\small
\centering
\caption{ININ name databases with baseline accuracies}
\label{tab:datasets}
\setlength{\tabcolsep}{2pt}
\footnotesize
\renewcommand{\arraystretch}{0.9}
\begin{tabular}{@{} *{5}{c} @{}} \toprule%
Database & Grammar & No. of Unique      & No. of Name           & NER \\
         & Size    & Names (Incorrect)  & Instances (Incorrect) & (\%) \\\midrule
phase 1 & 13875 & 12419 (5307) & 38806 (8083) & 20.83 \\
phase 2 & 13875 & 12662 (3998) & 42055 (6043) & 14.37 \\\bottomrule
\end{tabular}
\myfigureshrinker
\end{table}

Recognition is normally more challenging when the grammar size increases, since names are more dense in the pronunciation space and more easily confused with others. Here NER is evaluated with a growing grammar size $G$, and data in both phases were randomly segmented into subsets $\mathcal{C}_G$, with $G=1000,2000,\ldots,13000, 13875$, where the larger subset always includes the smaller one. 
%


\section{Overview of Grammar-based ASR}
\label{sec:overview}

\begin{figure}[htb]
 \centering
  \begin{tabular}{c}
 	\includegraphics[width = .4\textwidth]{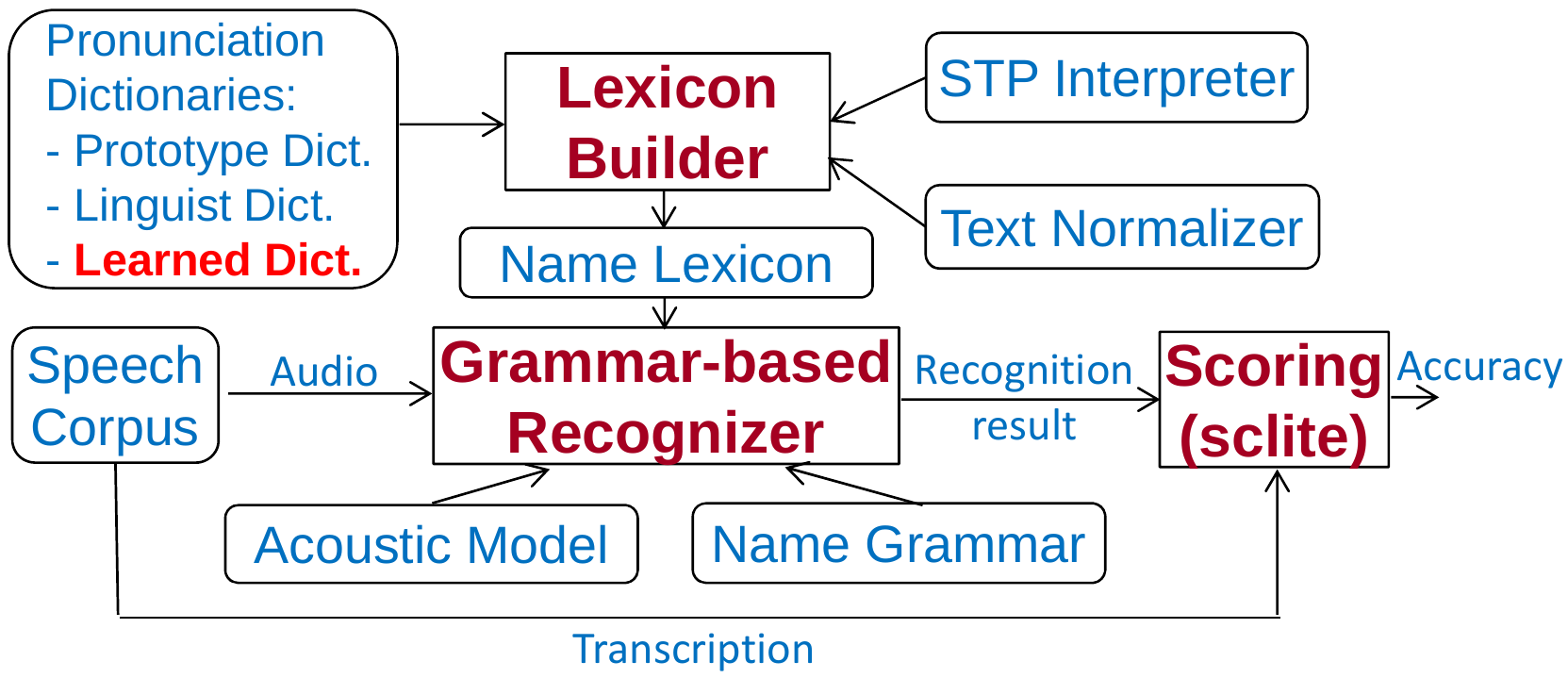}
  \end{tabular}
  \caption{Structure of grammar-based name recognition\label{fig:asr_diagram}}
\myfigureshrinker
\end{figure}

%
Grammar-based ASR is used to recognize input speech as one of the entries specified in the grammar file. For example, if the grammar contains various names, the input speech will be recognized as one of the most likely name or the system will report ``no match'', if it is not close to any name. This work used Interaction Speech Recognition\textsuperscript{\textregistered}, a grammar-based ASR developed at ININ. Fig. \ref{fig:asr_diagram} illustrates the main components with both acoustic and language resources. The acoustic information is modeled as Hidden Markov Model-Gaussian Mixture Model (HMM-GMM). The front end for this systems uses Mel-Frequency Cepstral Coefficients (MFCCs) transformed using Linear Discriminant Analysis (LDA). The language resource is provided as a name grammar according to the speech recognition grammar specification (SRGS) \cite{grammar}. The linguistic information is encoded using a lexicon containing text normalizers, pronunciation dictionaries, and a decision-tree-based spelling-to-pronunciation (STP) predictor. 
The work here updates the pronunciation dictionaries by adding learned pronunciations to build a better lexicon for recognition. 

%
%

\section{Preliminaries}
\label{sec:preliminary}

To better describe the pronunciation learning algorithm in Sec. \ref{sec:pronlearn}, this section introduces three preliminary interrelated concepts, including a) confusion matrix, b) pronunciation space and distance, and c) generation of candidate pronunciations. 

\subsection{Confusion Matrix}
\label{subsec:confusion_mtx}

This work used the \textit{Arpabet} phoneme set of 39 phonemes \cite{CMU_dict:2013} to construct a $39 \times 39$ confusion matrix $\mathcal{M}$. The value $\mathcal{M}(p_i,p_j)$ serves as a similarity measurement between phonemes $p_i$ and $p_j$. The smaller the value, the more similar they are. It considers both acoustic and linguistic similarities and is formulated as:
\begin{equation}
\small
\label{eq:confusion_mtx}
  \mathcal{M}(p_i,p_j) = \mathcal{M}_{\textrm{acoustic}}(p_i,p_j) \cdot \
  \mathcal{M}_{\textrm{linguistic}}(p_i,p_j)
\end{equation}
To construct $\mathcal{M}_\textrm{acoustic}$, phoneme alignment was performed on the Wall Street Journal (WSJ) corpus to find the average log-likelihood of recognizing phoneme $p_{i}$ as $p_{j}$. These values were then sign-flipped and normalized, so the diagonal values in $\mathcal{M}_\textrm{acoustic}$ are all zeros. $\mathcal{M}_\textrm{linguistic}$ is a symmetric binary matrix where $\mathcal{M}_\textrm{linguistic}(p_i,p_j)=0$ if $p_i$ and $p_j$ are in the same linguistic cluster. The confusion between $p_i$ and $p_j$ is linguistically likely even though they may acoustically sound very different, and vice versa. Tab. \ref{tab:phone_clusters} shows the 16 clusters defined by in-house linguists based on linguistic similarities. Using Eq. (\ref{eq:confusion_mtx}), the combined confusion matrix $\mathcal{M}$ prioritizes the linguistic similarity, where the acoustic similarity is considered only when the phonemes are not in the same linguistic cluster.

\begin{table}[htb!]
\centering
\caption{Linguistic phoneme clusters}
\label{tab:phone_clusters}
\setlength{\tabcolsep}{2pt}
\renewcommand{\arraystretch}{0.9}
\begin{small}
\begin{tabular}{@{} cl|cl|cl|cl @{}} \toprule
\textbf{1} & iy, ih, ay, y & \textbf{5} & ey, eh & \textbf{9} & ae, aa, ao, ah, aw & \textbf{13} & ow, oy \\
\textbf{2} & uw, uh, w & \textbf{6} & er, r, l & \textbf{10} & p, b & \textbf{14} & t, d \\
\textbf{3} & k, g & \textbf{7} & f, v & \textbf{11} & s, z, sh, zh & \textbf{15} & ch, jh \\
\textbf{4} &  m & \textbf{8} & n, ng & \textbf{12} & th, dh & \textbf{16} & hh \\
\bottomrule
\end{tabular}
\end{small}
\end{table}

\subsection{Pronunciation Space and Distance Measurement}
\label{subsec:pron_space}

Pronunciation space is spanned by all possible pronunciations (phoneme sequences). Sequences are considered as points in this space and the ``distances'' between them are computed using a confusion matrix $\mathcal{M}$. The distance $d(\mathcal{P}_i, \mathcal{P}_j)$  between two pronunciations $\mathcal{P}_i=[p_1,p_2,\ldots,p_M]$ and $\mathcal{P}_j=[q_1,q_2,\ldots,q_N]$, where $M,N$ are the lengths of phoneme sequences, is measured using Levenshtein distance with Dynamic Programming \cite{heeringa2004measuring}. It is then  normalized by the maximum length of these two, i.e., $d(\mathcal{P}_i, \mathcal{P}_j) = C(M,N)/\max \{M,N\}$.
For a database with grammar size $G$, a $G \times G$ name distance matrix $\mathcal{N}$ is pre-computed before pronunciation learning, where $\mathcal{N}(s,t)$ indicates the distance between name $\mathcal{N}_s$ and $\mathcal{N}_t$.

\subsection{Generation of Candidate Pronunciations}
\label{subsec:can_prons}

pronunciation learning of a target word $\mathcal{W}_t$ requires generating a pool of candidate pronunciations $\mathcal{P}_\textrm{Can}$ ``around'' the baseline pronunciation $\mathcal{P}_\textrm{Base}$ in the pronunciation space to search from. Given $\mathcal{P}_\textrm{Base}=[p_M,p_{M-1},\ldots,p_1]$, where $M$ is the length of $\mathcal{P}_\textrm{Base}$, by thresholding the $m^{\textrm{th}}$ phoneme $p_m$ in the confusion matrix $\mathcal{M}$ with search radius $r_m$, $m \in [1,M]$, one can find $N_m$ candidate phonemes (including $p_m$ itself, since $\mathcal{M}(p_m,p_m)=0 < r_m $), which can be indexed in the range $[0,1,\ldots,n_m,\ldots, N_{m-1}]$. Note that the phoneme symbol $p_m$ in $\mathcal{P}_\textrm{Base}$ are indexed in reverse order and the index of candidate phonemes $n_m$ for $p_m$ start from 0, rather than 1. This is intentional to make it easier to describe the candidate pronunciation indexing later in this section. Here we use the same search radius $r_0$ to search potential replacements for each phoneme, i.e., $r_0=r_1=\cdots=r_M$, where $r_0$ is experimentally determined by the workload (i.e. the number of misrecognized name instances) required for pronunciation learning.

After finding $N_m$ candidate phones for $p_m$ using search radius $r$, the total number of candidate pronunciations ($\mathcal{P}_\textrm{Can}$) $X$ can be calculated by $X = \prod_{m=1}^{M}N_m$.
%
%
For example, given the word {\tt paine} with $\mathcal{P}_\textrm{Base} = \textrm{[p ey n]}$, here $M=3$ and there are $N_3=2, N_2=4, N_1=2$ candidate phonemes for $p_m, m \in [1,M]$. The phoneme candidates for substitution are listed in Tab. \ref{tab:pron_example}, and Tab. \ref{tab:candidate_example} shows all 16 $\mathcal{P}_\textrm{Can}$ ($X=16$) with repetition patterns underlined. 
\begin{table}[tb]
\myfigureshrinker
\centering
\caption{phoneme substitution candidates for $\mathcal{P}_\textrm{Base}({\tt paine})$}
\label{tab:pron_example}
\setlength{\tabcolsep}{3pt}
\renewcommand{\arraystretch}{0.9}
\begin{footnotesize}
\begin{tabular}{@{} c|lll @{}} \toprule
$p_m$ in $\mathcal{P}_{\mathrm{Base}} (M=3)$ & $p_3=\textrm{p}$ & $p_2=\textrm{ey}$ & $p_1=\textrm{n}$ \\
Number of candidates $N_m$ & $N_3=2$ & $N_2=4$ & $N_1=2$ \\
Candidate index $n_m$ & $n_3 \in [0,1]$ & $n_2 \in [0,3]$ & $n_1 \in [0,1]$ \\\midrule
\multirow{1}*{Phoneme} & b (0) & eh (0), ey (1) & n (0) \\
candidate ($n_m$) & p (1) & iy (2), ih (3) & ng (1) \\\bottomrule
\end{tabular}
\end{footnotesize}
\myfigureshrinker
\end{table}
\begin{table}[tb]
\centering
\caption{Candidate pronunciations of the word {\tt paine} with their pronunciation and phoneme indices}
\label{tab:candidate_example}
\setlength\extrarowheight{0pt}
\setlength{\tabcolsep}{3pt}
\renewcommand{\arraystretch}{-1}
\small
\begin{footnotesize}
\begin{tabular}{@{} *{7}c*{7}c @{}} \toprule
$x$ & $n_3$ & $n_2$ & $n_1$ & \multicolumn{3}{c}{$\mathcal{P}_{\textrm{Can}}$} & \
$x$ & $n_3$ & $n_2$ & $n_1$ & \multicolumn{3}{c}{$\mathcal{P}_{\textrm{Can}}$} \\\midrule
$\textbf{0}$ & $0$ & $0$ & $0$ & b & eh & n & \
$\textbf{8}$ & $1$ & $0$ & $0$ & p & eh & n \\
\cmidrule(l){4-4} \cmidrule(l){7-7} \cmidrule(l){11-11} \cmidrule(l){14-14}
$\textbf{1}$ & $0$ & $0$ & $1$ & b & eh & ng & \
$\textbf{9}$ & $1$ & $0$ & $1$ & p & eh & ng \\
\cmidrule(l){3-4} \cmidrule(l){6-7} \cmidrule(l){10-11} \cmidrule(l){13-14}
$\textbf{2}$  & $0$ & $1$ & $0$ & b & ey & n & \
$\textbf{10}$ & $1$ & $1$ & $0$ & p & ey & n \\
\cmidrule(l){4-4} \cmidrule(l){7-7} \cmidrule(l){11-11} \cmidrule(l){14-14}
$\textbf{3}$  & $0$ & $1$ & $1$ & b & ey & ng & \
$\textbf{11}$ & $1$ & $1$ & $1$ & p & ey & ng \\
\cmidrule(l){3-4} \cmidrule(l){6-7} \cmidrule(l){10-11} \cmidrule(l){13-14}
$\textbf{4}$  & $0$ & $2$ & $0$ & b & iy & n & \
$\textbf{12}$ & $1$ & $2$ & $0$ & p & iy & n \\
\cmidrule(l){4-4} \cmidrule(l){7-7} \cmidrule(l){11-11} \cmidrule(l){14-14}
$\textbf{5}$  & $0$ & $2$ & $1$ & b & iy & ng & \
$\textbf{13}$ & $1$ & $2$ & $1$ & p & iy & ng \\
\cmidrule(l){3-4} \cmidrule(l){6-7} \cmidrule(l){10-11} \cmidrule(l){13-14}
$\textbf{6}$  & $0$ & $3$ & $0$ & b & ih & n & \
$\textbf{14}$ & $1$ & $3$ & $0$ & p & ih & n \\
\cmidrule(l){4-4} \cmidrule(l){7-7} \cmidrule(l){11-11} \cmidrule(l){14-14}
$\textbf{7}$  & $0$ & $3$ & $1$ & b & ih & ng & \
$\textbf{15}$ & $1$ & $3$ & $1$ & p & ih & ng \\
\cmidrule(l){2-4} \cmidrule(l){5-7} \cmidrule(l){9-11} \cmidrule(l){12-14}
\end{tabular}
\end{footnotesize}
\myfigureshrinker
\end{table}
Meanwhile, the distance from $\mathcal{P}_\textrm{Base}$ of $\mathcal{W}_t$ to the farthest candidate pronunciation is defined as its outreach distance $d_t$, which is later used to define the scope in finding $\mathcal{W}_t$'s neighboring words. It is formulated below:
%
%
\begin{eqnarray}
\small
\begin{aligned}
\label{eq:outreach_dist}
d_t &= \frac{1}{M}\sum_{m=1}^{M} d_t(p_m) \\
    &= \frac{1}{M}\sum_{m=1}^{M} \max_{n_m \in [0, N_{m}-1]} \mathcal{M}(p_m,p_{m}(n_m)) ,
\end{aligned}    
\end{eqnarray}
and $p_{m}(n_m)$ is the $n_m^{\textrm{th}}$ candidate phoneme alternative for $p_m$. 

After generating candidate pronunciation list $\mathcal{P}_\textrm{Can}$ from $\mathcal{P}_\textrm{Base}$ using this method, fast one-to-one mapping/indexing between phoneme indices $(n_M, n_{M-1}, \ldots, n_m, \ldots, n_1)$ and pronunciation index $x$, $x \in[0, X-1]$ is essential for efficient candidate pronunciation lookup and pronunciation list segmentation based on the phoneme position index $m$ during pronunciation learning. Therefore, a pair of bi-directional mapping functions is provided in Eq. (\ref{eq:phonei2proni}) and Eq. (\ref{eq:proni2phonei}). For example, [p iy ng] can be indexed by both $\mathcal{P}_\textrm{Can}(x=13)$ and $\mathcal{P}_\textrm{Can}(n_3=1, n_2=2, n_1=1)$.
%
%
\begin{equation}
\small
\label{eq:phonei2proni}
x = \sum_{m=1}^{M}n_m(\prod_{i=0}^{m-1}N_{i}), n_m\in[0,N_{m-1}], N_0 = 1,
\end{equation}
%
%
\begin{equation}
\small
\label{eq:proni2phonei}
n_m = \left[\dfrac{x - x\bmod(\prod_{i=0}^{m-1}N_{i})}{\prod_{i=0}^{m-1}N_{i}}\right] \bmod N_m, N_0 = 1.
\end{equation}

The example shown above illustrates candidate pronunciation generation with phoneme replacement. This method can be easily extended to include more candidates with phoneme deletion, by introducing a special ``void'' phoneme. However, it does not handle phoneme insertion since it may include too many possible candidates.

\section{Pronunciation Learning Algorithm}
\label{sec:pronlearn}

Pronunciation learning aims to find better alternative pronunciation for misrecognized names through a pronunciation generation and pruning process, which maximizes the accuracy improvement on a regional nameset $\mathcal{D}_r$including the target name $\mathcal{N}_t$ and its nearby similar names $\mathcal{N}_c$. The learning is performed for all misrecognized names. However, it is only applied on a word basis, to the misrecognized words in the misrecognized names. The following subsections first introduce the main word pronunciation learning algorithm and then elaborate on the key components. 

\subsection{Algorithm Outline}
\label{subsec:algorithm}

\begin{enumerate}[1.]\itemsep0pt
\item Set phoneme search radius $r_0$, and upper bounds on the number of total pronunciations per name $K_1$ and per word $K_2$.
\item Perform baseline name recognition and collect all misrecognized name instances in $\mathcal{D}_e$.
\item For each target name $\mathcal{N}_t$ with error instances in $\mathcal{D}_e$:
  \begin{enumerate}[a.]\itemsep0pt
  \item Compute its $\mathcal{P}_\textrm{Can}$ with $r_0$ and outreach distance $d_t$ in Eq. (\ref{eq:outreach_dist}) to find its corresponding regional nameset $\mathcal{D}_r$.
  \item For each misrecognized word instance $\mathcal{W}_{t}(i)$, find the best pronunciation $\mathcal{P}^{*}(i)$ using \textbf{hierarchical pronunciation determination} and get the accuracy increment $\tilde{A}(i)$ on $\mathcal{D}_r$ by adding $\mathcal{P}^{*}(i)$ into dictionary. 
  \item Sort $\mathcal{P}^{*}(i)$ by $\tilde{A}(i)$ and keep up to $K_1$ pronunciations in $\mathcal{P}_\textrm{Learned}$ dictionary.	
  \end{enumerate}
\item For each target word $\mathcal{W}_t$ with learned pronunciations:
	\begin{enumerate}[a.]\itemsep0pt
	\item Find all names containing $\mathcal{W}_t$ to form a nameset $\mathcal{D}_{w}$.
	\item Evaluate $\mathcal{P}_\textrm{Learned}(\mathcal{W}_t)$ significance by their accuracy boost $\tilde{A}_{w}$ on $\mathcal{D}_{w}$ and keep up to top $K_2$ pronunciations.
	\end{enumerate}
\item Combine all $\mathcal{P}_\textrm{Learned}$ for error words $\mathcal{W}_e$ after pruning and replace $\mathcal{P}_\textrm{Base}(\mathcal{W}_e)$ with $\mathcal{P}_\textrm{Learned}(\mathcal{W}_e)$ in dictionary. 		   
\end{enumerate}     

\subsection{Hierarchical Pronunciation Determination}
\label{subsec: pron_determination}

\begin{figure}[htb!]
 \centering
 	\includegraphics[height=2.25cm, width=7cm]{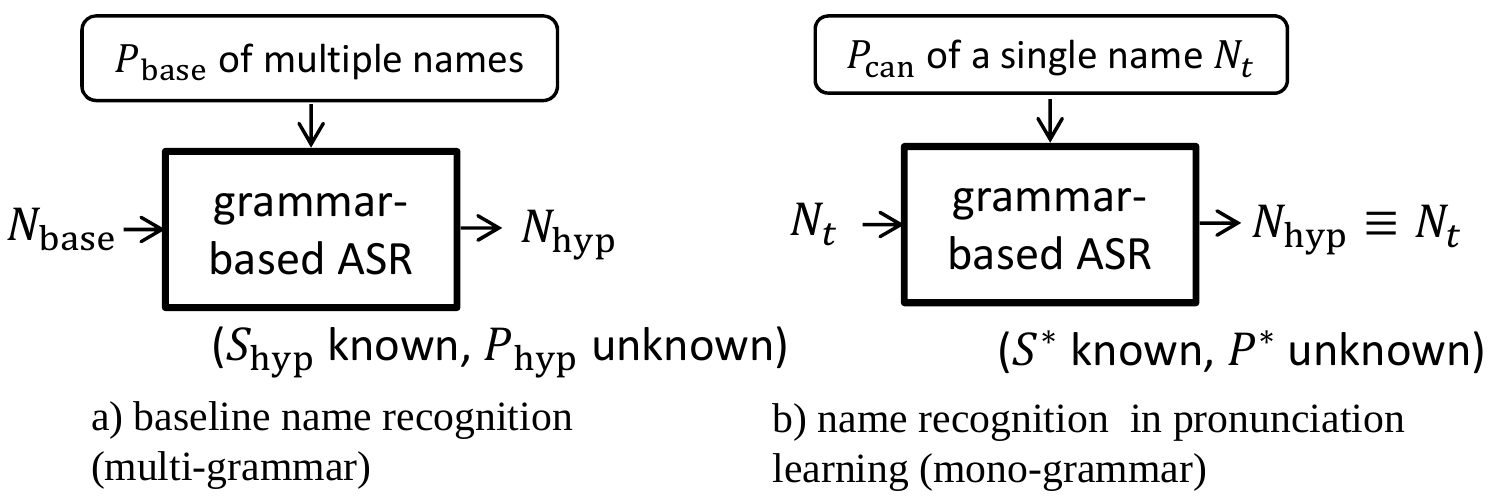}
    \caption{Simplified demos of multi-gram and mono-gram ASR \label{fig:hpd_demo}}
\myfigureshrinker    
\end{figure}

Generally, given an input test name $\mathcal{N}_\textrm{Base}$ to the grammar-based ASR, it outputs a hypothesized name $\mathcal{N}_\textrm{Hyp}$ associated with the highest hypothesized score $S_\textrm{Hyp}$. However, if $\mathcal{N}_\textrm{Hyp}$ has multiple pronunciations, which one is actually used to yield $S_\textrm{Hyp}$ is not provided for decoding efficiency (Fig. \ref{fig:hpd_demo}a). It is similar in the case of pronunciation learning (Fig. \ref{fig:hpd_demo}b). By providing massive number of $\mathcal{P}_\textrm{Can}$ for an ASR with single grammar (grammar contains only one name $\mathcal{N}_{t}$), only the highest hypothesized score $S^{*}$ is yielded and the associated best pronunciation $\mathcal{P}^*$ is not provided. In order to find $\mathcal{P}^*$ from $\mathcal{P}_\textrm{Can}$, hierarchical pronunciation determination with $\mathcal{P}_\textrm{Can}$ segmentation is used, by determining its phoneme one at a time.
%
For simplicity, an example to determine $\mathcal{P}^*$ for the word {\tt paine} is demonstrated. The same method applies to a name (concatenation of words) as well. 
\begin{figure}[tb]
 \centering
 	\includegraphics[height=5cm, width=7cm]{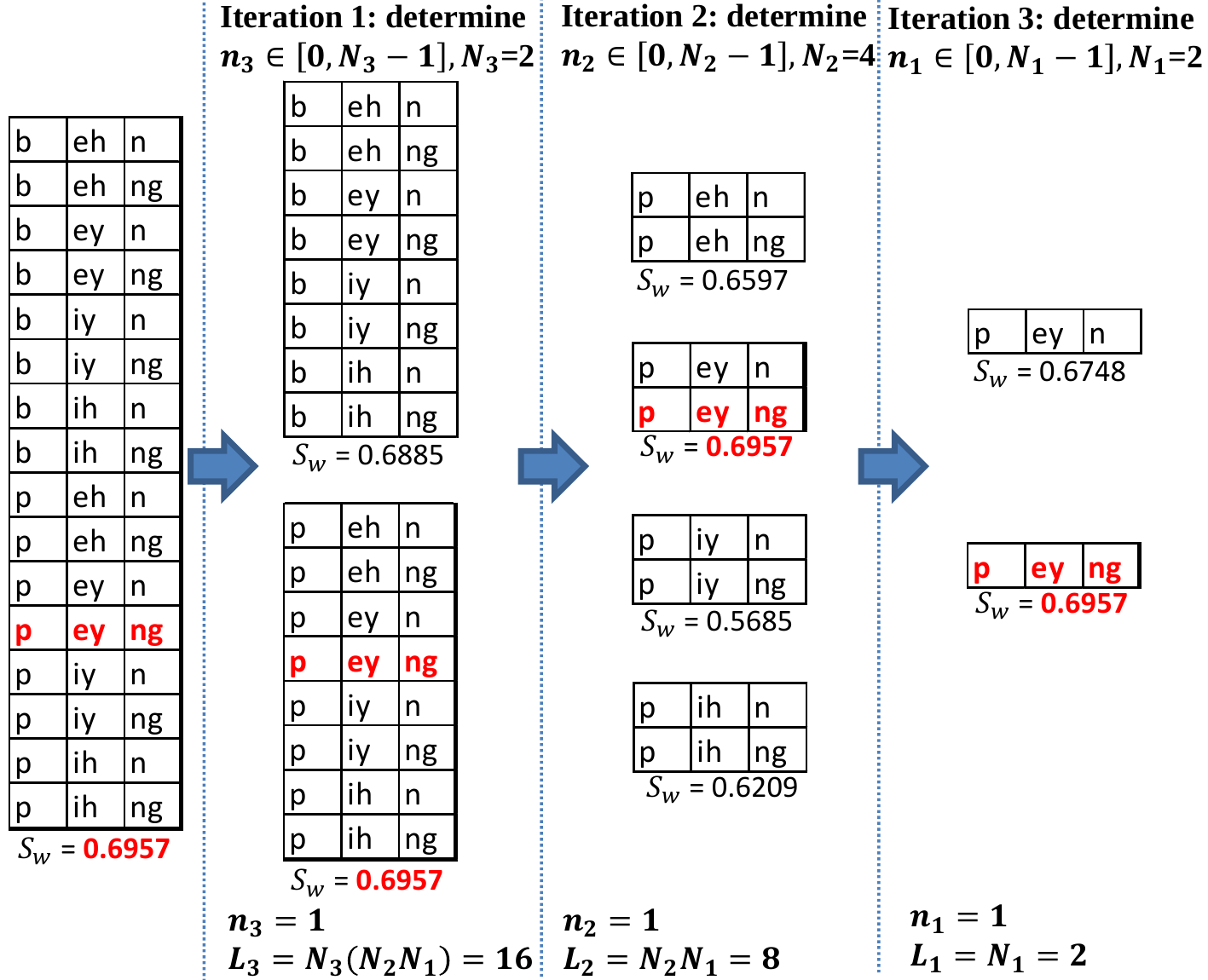}
    \caption{Hierachical pronunciation determination on the word {\tt paine}\label{fig:hpd_example}}
\myfigureshrinker    
\end{figure}
In Fig. \ref{fig:hpd_example}, the phonemes in $\mathcal{P}^*$ are determined in the order of $p_3^*=\textrm{p}$ $\rightarrow$ $p_2^*=\textrm{ey}$ $\rightarrow$ $p_1^*=\textrm{ng}$, by tracking the $\mathcal{P}_\textrm{Can}$ segmentation with highest confidence score $S_\mathcal{W}=0.6957$.

In general, given $N_i$ phoneme candidates for the $i^{\textrm{th}}$ phoneme of $\mathcal{P}_\textrm{Base}$, $L_m=\prod_{i=1}^{m}N_i$ is the number of pronunciations processed to determine the $m^{\textrm{th}}$ phoneme, and $L = \sum_{m=1}^{M}L_m$ is the total number of pronunciations processed while learning the pronunciation for one word. In addition, the number of times running the recognizer is $\sum_{m=1}^{M}N_m$. Given that the computational cost of running the recognizer once is $T_1$ and processing each candidate pronunciation is $T_2$, where $T_1 \gg T_2$, the total computational cost of running hierarchical pronunciation determination $T$ is approximately
\begin{eqnarray}
\small
\begin{aligned}
\label{eq:compcost}
T &\approx T_{\textrm{Run}} + T_{\textrm{Pron}} \\
  &= \left(\sum_{m=1}^{M}N_{m}\right)T_1 + \left(\sum_{m=1}^{M} \left( \prod_{i=1}^{m}N_i \right) \right)T_2.
\end{aligned}
\end{eqnarray}
For example, when determining phonemes in the order of $p_3 \rightarrow p_2 \rightarrow p_1 $ (natural order) in Figure \ref{fig:hpd_example}. 
%
\begin{eqnarray}
\small
\begin{aligned}
\label{eq:compcost_example}
T_{\tt paine} &\approx (2+4+2)T_1 \quad + \left[(2 \cdot 4 \cdot 2) + (4 \cdot 2) + 2 \right] T_2 \\
  &= 8T_1 + 26T_2 .  
\end{aligned}
\end{eqnarray}

Since $T_1 \gg T_2$, $T$ is mainly determined by $T_{\textrm{Run}}$, i.e. the factor $\sum_{m=1}^{M}N_{m}$. Comparing with the brute-force method of evaluating candidate pronunciations one-by-one, associated with the factor $\prod_{m=1}^{M}N_{m}$, this algorithm is significantly faster.



\section{Optimization in Implementation}
\label{sec:improvement}


\subsection{Search Radius Reduction}

In the step 3b of Sec. \ref{subsec:algorithm}, if too many alternatives are generated for a particular word, due to the phoneme sequence length $M > M_\mathrm{max}$, search radius reduction is triggered to reduce the computational cost by decreasing the phoneme search radius form $r_0$ to $\frac{M_\mathrm{max}-1}{M-1}r_0$. For example, the word {\tt desjardins} with $\mathcal{P}_\textrm{Base}=\textrm{[d eh s zh aa r d iy n z]}$ is a long word with $M=10$, and phonemes \{eh, s, zh , aa, iy, z\} have more than 5 phoneme candidates each. The total number of $\mathcal{P}_\textrm{Can}$ is $4,536,000$ which requires much longer learning time than regular words. There are less than 20\% of words in $\mathcal{D}_\textrm{Train}$ that triggered this. However, the average word pronunciation length was reduced from 20,204 to 11,941. Both $r_0$ and $M_{\textrm{max}}$ in are determined experimentally, here $r_0=3$ and $M_{\textrm{max}}=6$. This method narrows pronunciation variants search to the more similar ones. 

\subsection{Phoneme Determination Order Optimization}

\begin{figure}[htb]
 \centering
 	\includegraphics[height=5cm, width=7cm]{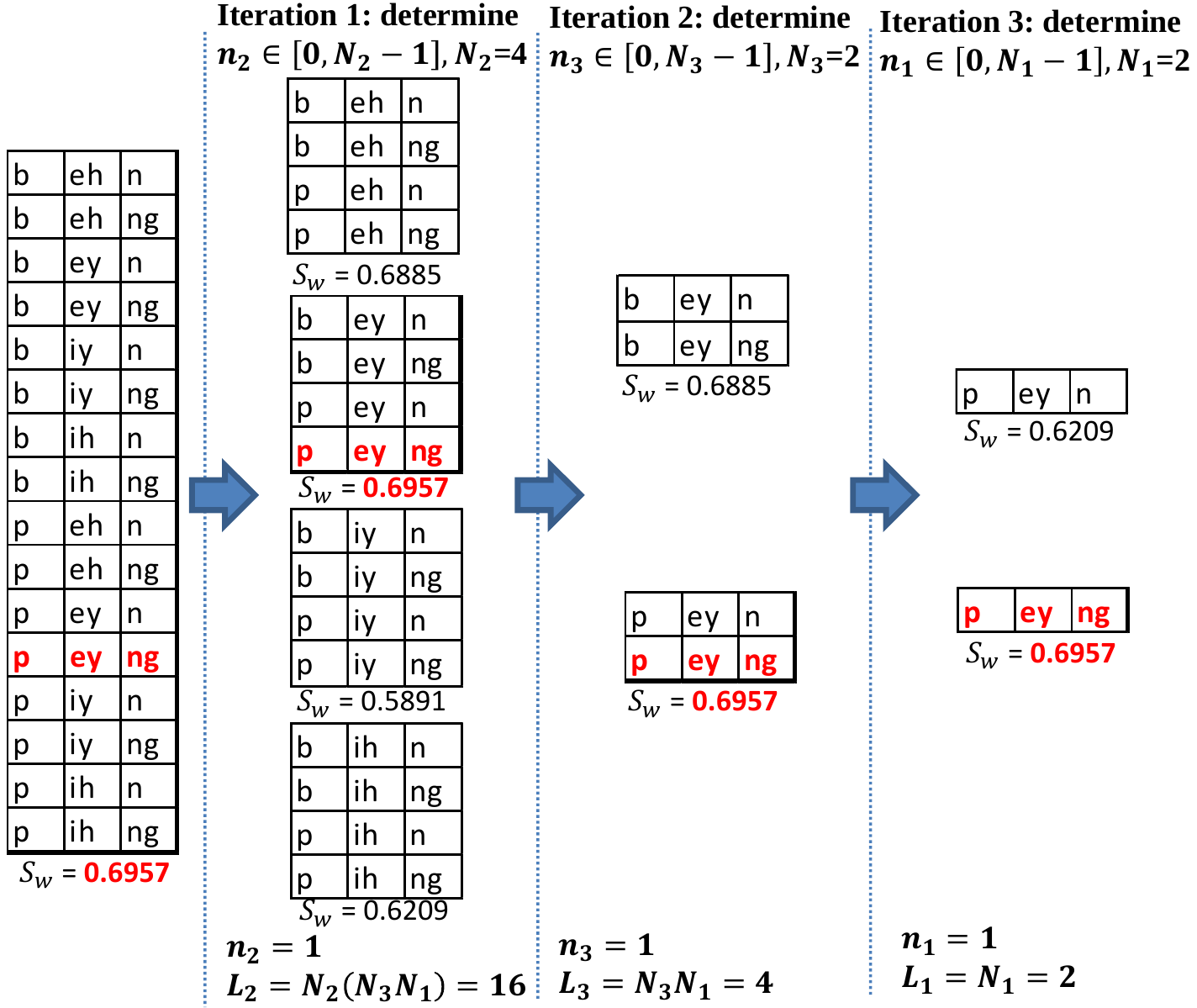}
    \caption{Hierachical pronunciation determination on the word {\tt paine}\label{fig:hpd_descend} with phonemes determined in descending order by number of candidates ($N_2=4 \rightarrow N_3=2 \rightarrow N_1=2$)}
\myfigureshrinker
\end{figure}

Given $\{N_M,N_{M-1},\ldots,N_1\}$ are the number of phoneme candidates for phonemes $\{p_M,p_{M-1},\ldots,p_1\}$ in $\mathcal{P}_\textrm{Base}$. Fig. \ref{fig:hpd_example} shows that the phonemes are determined in the natural order of $N_m$, such as $p_3 \rightarrow p_2 \rightarrow p_1$, and the total number of $\mathcal{P}_\textrm{Can}$ processed is $L_{\textrm{Natural}} = 26$. However, if they are determined in the descending order of $N_m$, such as $p_2 \rightarrow p_3 \rightarrow p_1$ ($N_2=4 \geq N_3=2 \geq N_1=2$), then the number of $\mathcal{P}_\textrm{Can}$ processed is minimized as $L_{\textrm{Descend}}=22 < L_{\textrm{Natural}}=26$ (Fig. \ref{fig:hpd_descend}). Generally, it can be mathematically proven that $L_\textrm{Descend} \leq L_\textrm{Natural} \leq L_\textrm{Ascend}.$

\section{Results and Conclusions}
\label{sec:results}

\begin{figure}[htb]
\myfigureshrinker
 \centering
 	\includegraphics[height=2.5cm, width=7.5cm]{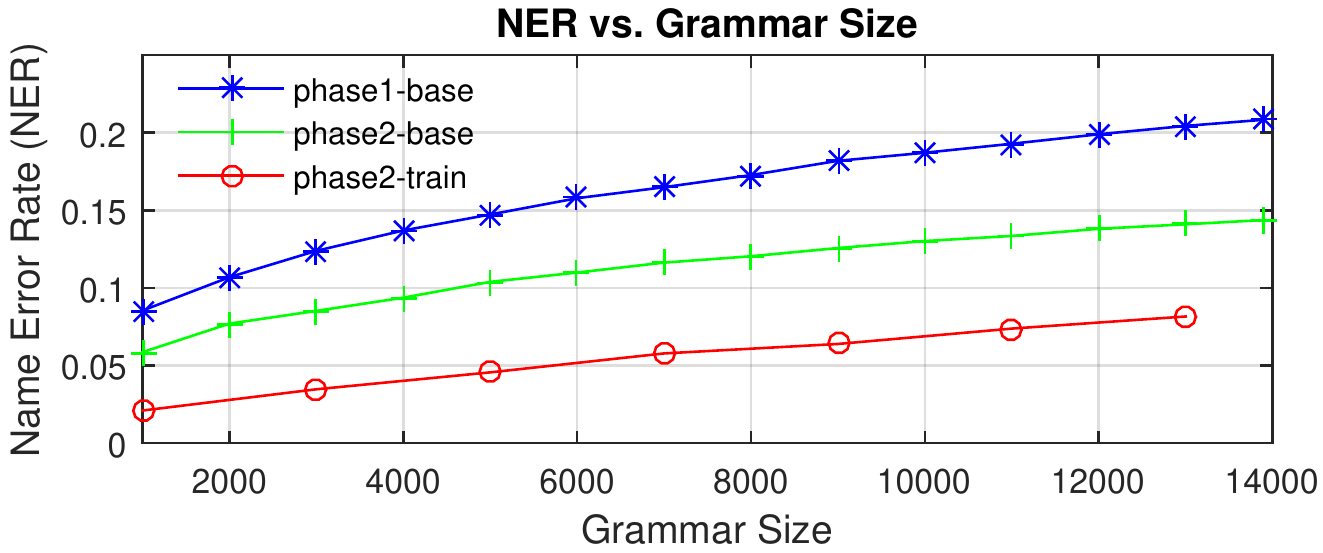}
    \caption{NERs before and after pronunciation learning\label{fig:ner}}
\myfigureshrinker
\end{figure}

The improvement from baseline varies from the experimental settings, e.g. 1) how challenging the database is (percentage of uncommon words); 2) the dataset and grammar sizes; 3) the quality of audio recording and 4) the ASR acoustic modeling, etc. The namesets in Tab. \ref{tab:datasets} contain 13.4\% uncommon words (4.6\% native, 8.8\% non-native). Tab. \ref{tab:results} and Fig. \ref{fig:ner} show the baselines are already with competitive accuracies, since the dictionaries provide canonical pronunciations of 97\%+ words, and the rest are generated from a well-trained STP with perfect-match accuracy 55\% on 5800 words reserved for testing.     
%
%
The pronunciations learned from phase 1 update the lexicon and are tested in phase 2.  All NERs grow when $G$ increases, and $\textrm{NER}_\textrm{learn}^{(2)}$ is much lower but grows slightly faster than the other two.

Beaufays et al. \cite{beaufays2003learning} achieved ERR 40\% with 1600 names, compared to a baseline letter-to-phone pronunciation engine. We obtained similar ERR with a much larger grammar size (42.13\% with 13000 names) with a much better baseline. Compared with Reveil et al. \cite{reveil2012improving}, whose ERR was close to 40\% with 3540 names spoken by speakers from 5 different language origins, our dataset may not have such diversity but we achieved much higher ERR of around 58\% for a similar grammar size.   

%
%

\begin{table}[htb]
\myfigureshrinker
\captionsetup[table]{aboveskip=0pt}
\centering
\caption{Name Error Rate (NERs) and Error Reduction Rate (ERR)}
\label{tab:results}
\setlength{\tabcolsep}{1pt}
\footnotesize
\renewcommand{\arraystretch}{0.9}
\begin{tabular}{@{} *{8}{c} @{}} \toprule%
Grammar Size ($G$) & 1000 & 3000 & 5000 & 7000 & 9000 & 11000 & 13000 \\\midrule
phase 1 base ($\textrm{NER}_{\textrm{base}}^{(1)}$) & 8.54 & 12.39 & 14.73 & 16.49 & 18.19 & 19.29 & 20.44 \\
phase 2 base ($\textrm{NER}_{\textrm{base}}^{(2)}$) & 5.86 & 8.52 & 10.39 & 11.63 & 12.57 & 13.35 & 14.10 \\
phase 2 learn ($\textrm{NER}_{\textrm{learn}}^{(2)}$) & 2.10 & 3.47 & 4.56 & 5.78 & 6.39 & 7.38 & 8.16 \\\midrule
phase 2 ERR & 64.16 & 59.27 & 56.11 & 50.30 & 49.16 & 44.72 & 42.13\\\bottomrule
\end{tabular}
\myfigureshrinker
\end{table}

This pronunciation learning algorithm is an essential complement to a) STP/G2P interpreters and b) content-adapted pronunciation dictionaries made by linguists. When these two are not sophisticated enough to cover pronunciation variations, it can help to build a better lexicon, and ASR accuracy can significantly be improved, especially when the grammar size is not too large. As indicated in Tab. \ref{tab:results}, ERR of phase 2 tends to decrease when the grammar size increases, since there is not much room for learning when one name is surrounded by many other names with similar pronunciations. Similarly, the learned dictionary is also dependent on grammar size, i.e., one dictionary learned from a small database might not be a good fit for a much larger database, since it may be too aggressive in learning, while a larger database requires a more conservative approach to learn.
In the future, learned pronunciations can be used to improve the STP interpreter by generating alternative spelling-to-pronunciation interpretation rules, so it can automatically output alternative pronunciations covering new names, and provide a baseline that is good enough even without learning. 


  \newpage
  \eightpt
  \bibliographystyle{IEEEtran}

  \bibliography{references}


\end{document}